\documentclass[conference]{IEEEtran}
\IEEEoverridecommandlockouts
\usepackage{comment}
\usepackage{color}
\usepackage{algorithm} 
\usepackage{algpseudocode} 
\usepackage[table,xcdraw]{xcolor}
\usepackage{cite}
\usepackage{amsmath,amssymb,amsfonts}
\usepackage{graphicx}
\usepackage{textcomp}
\usepackage{xcolor}
\usepackage{multirow}
\usepackage{amsmath}
\usepackage{color}
\usepackage[table,xcdraw]{xcolor}
\usepackage{algorithm} 
\usepackage{algpseudocode} 
\usepackage{subcaption}

\def\BibTeX{{\rm B\kern-.05em{\sc i\kern-.025em b}\kern-.08em
    T\kern-.1667em\lower.7ex\hbox{E}\kern-.125emX}}
\begin{document}

\title{SAPAG: A Self-Adaptive Privacy Attack From Gradients }

\author{
Yijue Wang\textsuperscript{1}, Jieren Deng\textsuperscript{1}, Dan Guo\textsuperscript{2}, Chenghong Wang\textsuperscript{3}, Xianrui Meng\textsuperscript{4}, Hang Liu\textsuperscript{5},\\ Caiwen Ding\textsuperscript{1},  Sanguthevar Rajasekaran\textsuperscript{1} \\
\textsuperscript{1}University of Connecticut, \textsuperscript{2}Northeastern University\\
\textsuperscript{3}Duke University\\
\textsuperscript{4}Amazon Web Services, Inc. \\
\textsuperscript{5}Stevens Institute of Technology \\
{\tt\small \{yijue.wang, jieren.deng, caiwen.ding, sanguthevar.rajasekaran\}@uconn.edu} \\
{\tt\small \{guo.dan\}@husky.neu.edu}\\
{\tt\small \{cw374\}@duke.edu}\\
{\tt\small \{xianruimeng\}@gmail.com}\\
{\tt\small \{hang.liu\}@stevens.edu}
}

\maketitle
\begin{abstract}
Distributed learning such as federated learning or collaborative learning enables model training on decentralized data from users and only collects local gradients, where data is processed close to its sources for data privacy.  The nature of not centralizing the training data addresses the privacy issue of privacy-sensitive data. Recent studies show that a third party can reconstruct the true training data in the distributed machine learning system through the publicly-shared gradients.
However, existing reconstruction attack frameworks lack generalizability on different Deep Neural Network (DNN) architectures and different weight distribution initialization, and can only succeed in the early training phase.
To address these limitations, in this paper, we propose a more general privacy attack from gradient, SAPAG, which uses a Gaussian kernel based of gradient difference as a distance measure. Our experiments demonstrate that SAPAG can construct the training data on different DNNs with different weight initializations and on DNNs in any training phases. 

\end{abstract}

\section{Introduction}

Distributed learning such as federated learning~ or collaborative learning\cite{chilimbi2014project, shokri2015privacy,moritz2015sparknet,zinkevich2010parallelized} refers to a setting where learning is done by multiple processors that are distributed (in space). It enables the devices at geographically different locations to collaboratively learn a machine learning model without sharing the local training data. 

On the one hand, the nature of not centralizing the training data on one server can help fast training on large-scale datasets. On the other hand, it addresses the privacy issue of privacy-sensitive data, such as personal health data\cite{jochems2016distributed}, genomic data, and confidential data in commercial entities. Federated learning works in a way that each local device can access the model parameters and train the model on the local training data, then only share the gradients back with the server. The server updates the shared model using the averages of gradients sent by multiple devices.

It is usually assumed that distributed learning is privacy guaranteed since the shared gradients contain no sufficient information to recover the original training data. However, several recent studies have made people rethink the privacy issue of distributed learning and federated learning by showing that it is possible to recover the training data just using the gradients. Researches such as~\cite{fredrikson2015model, hitaj2017deep, melis2019exploiting} show that 
the training data can be inferred from the gradients.
The DLG algorithm proposed by~\cite{zhu2019deep} extends the reconstruction of images to pixel-pixel accuracy by matching the gradients of a dummy image to the gradients of the attack target. 
However, 
existing works have at least one of the following limitations: (i) lack of generalizability on different Deep Neural Network (DNN) architectures; (ii) lack of generalizability on different weight distribution initialization; (iii) they only succeeded in the early training phase.

To address these issues, we provide a more general privacy attack
from gradients in this work: it can successfully reconstruct the training data on networks of any weights initialization or during training stage. The proposed method uses a Gaussian kernel of gradients differences as a basic measure of gradients distances, and the scaling factor in the Gaussian kernel is adaptive to the unique distribution of the gradients of the attack target. 
Our contributions are:
\begin{itemize}
\item {\bf{(Self-adaptive Attack)}} We develop a self-adaptive privacy attack from gradients algorithm (SAPAG) to reconstruct private training data from gradients in distributed learning systems. By self-learning the distribution of gradients, SAPAG can adapt to different weight distributions of Deep Neural Networks (DNNs).

\item{(\bf{Environment Generality})} SAPAG is a reconstruction attack framework that is compatible with various DNNs architectures (e.g. ResNet~\cite{he2016deep}, Transformer~\cite{vaswani2017attention}) with any kinds of weight distributions.

\item{(\bf{Effectiveness})} Our algorithm SAPAG can reconstruct the training data in any training phases of the DNN model.

\end{itemize}

Evaluations on different datasets and different DNNs show that SAPAG can reconstruct the training data on DNNs with different weight initializations and on DNNs in different training stages. In addition, SAPAG has higher reconstruction accuracy and faster convergence speed than the existing method, such as DLG. We also apply our attack on a transformer-based language model, on which it can also recover the token-wise training text. Studying the attack mechanism is beneficial to guide the designing of secure training schemes. Our attack can provide valuable information to the defense strategies in distributed learning.

\begin{figure}[t]
\centering
	\includegraphics[width=0.47\textwidth]{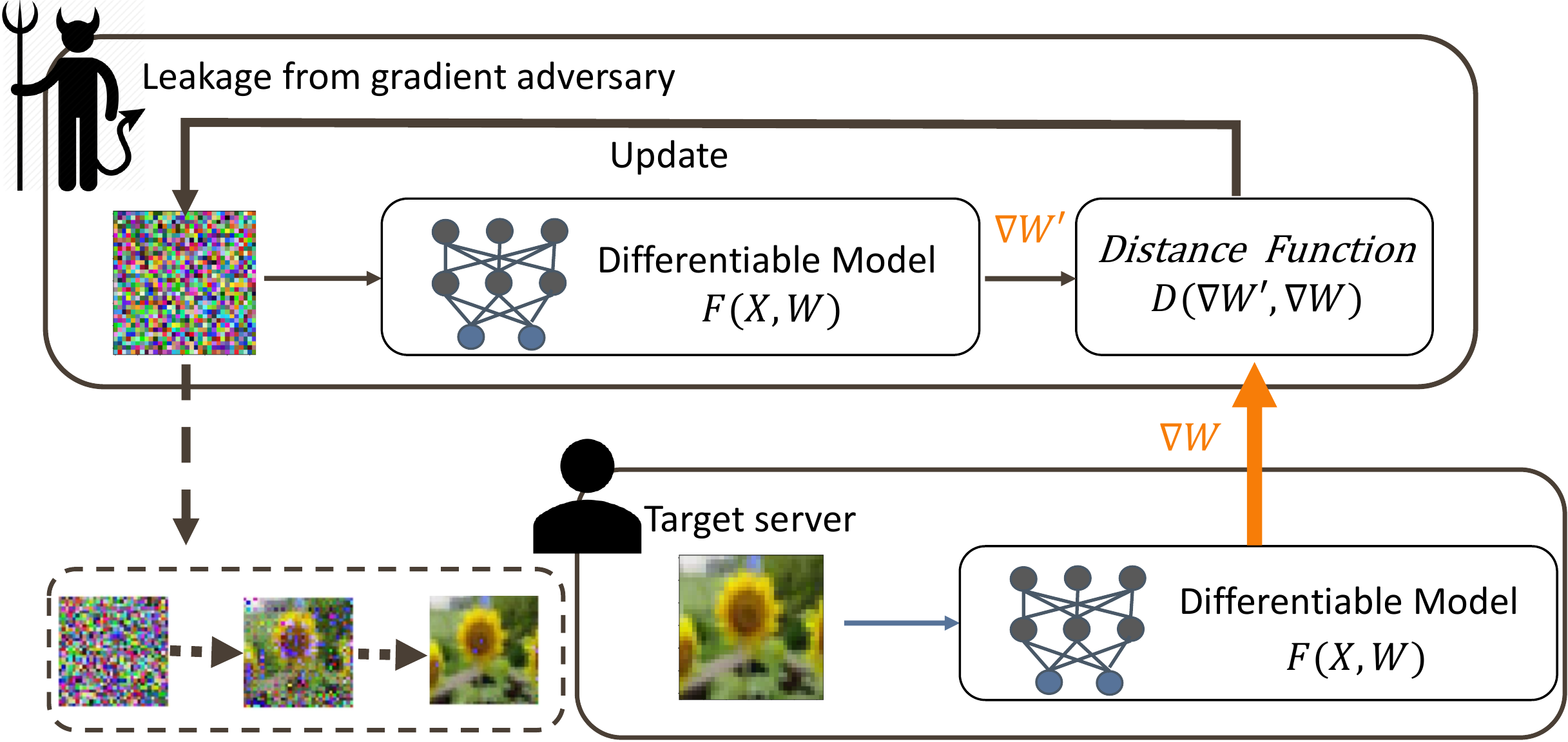}
    \vspace{-0.2 cm}
    \caption{ Privacy leakage from the gradient process.}
    \label{fig:intro}
\end{figure}



\section{Related Work}
\subsection{Distributed learning}
With the increasing size of training data and growing concerns on data privacy, training machine learning models efficiently and preserving training data privacy become challenging. Distributed learning (especially federated learning) has been developed to overcome this challenge \cite{akiba2017chainermn, dean2012large}. Instead of uploading all the data to a centralized server and training it jointly, distributed learning enables training on a large corpus of decentralized data on edge devices and only collects the local models or gradients for global synchronization on a central server~\cite{mcmahan2017federated, bonawitz2019towards,hard2018federated}.

\subsection{Privacy leakage from gradients}
Privacy leakage is studied in the training phase and prediction phase. Privacy attack from gradient and model inversion (MI) attack \cite{fredrikson2015model} aims at the training phase by constructing the features of the training data by using the correlation between the training data and the model output. \cite{fredrikson2015model} showed that it is possible to infer individual genomic data via access to a linear model for personalized medicine. Recent works extend MI attack to recover features of training data of DNN models. Privacy attack from gradients is different from previous MI attack. It reconstructs the training data exploiting their gradients in a machine learning model. The process of privacy leakage from gradients is shown at Figure~\ref{fig:intro}.   Although distributed learning system protects privacy by not sharing training data, research works have shown that it is possible to infer the information of training data from the shared gradients in both language tasks and computer vision tasks. ~\cite{melis2019exploiting} enables the identification of words used in the training tokens by analyzing the gradients of the embedding layer. \cite{goodfellow2014generative} proposes an attack algorithm to synthesize images mimicking the real training images by Generative Adversary Network (GAN) models.  
Besides the works that recover certain properties of the training data, a more recent work DLG~\cite{zhu2019deep} shows that it is possible to recover training data with pixel-wise accuracy for images and token-wise matching for texts by gradient matching. It first randomly generates a dummy image and a dummy label and then calculates the gradients according to the current weights of the network, and the dummy image and dummy label are updated by minimizing the Euclidean distance of gradients from the dummy image and the real training data.

\begin{figure}[ht]
\centering
	\includegraphics[width=0.35\textwidth]{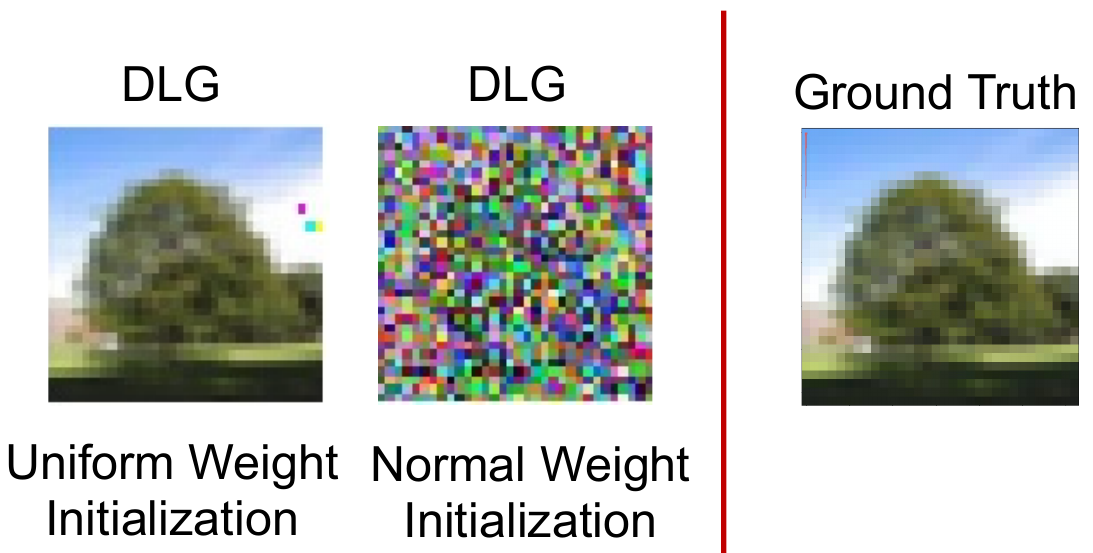}
    \vspace{-0.2 cm}
    \caption{ DLG's reconstructions of an image from CIFAR-100 for LeNet-5 with uniform and normal weight initialization}
    \label{fig:intro2}
\end{figure}

DLG achieves the reconstruction of images from different datasets on LeNet-5. However,  DLG has limitations on evaluating the performance thoroughly on different weight initialization settings, various networks, and different training stages (pretrained versus initialized). From our experiments, we have inferred that the DLG method is sensitive to the weight distribution of DNN and can only recover images under the uniform weight initialization, but cannot recover images under a normal weight initialization (as shown in Figure \ref{fig:intro2}) or for pretrained DNN models.

\section{Problem Statement}
We assume that an adversary cannot access the private data $({\mathbf{X}}, {\mathbf{Y}})$ in local training directly, but it is able to gain the gradients that the local devices shared and the current global model $\mathcal{F}(\mathbf{W},\mathbf{X})$.

The objective of the attack is to reconstruct the training data using the shared gradients. The reconstruction is pixel-pixel reconstruction for image data and token-token reconstruction for text data. Moreover, the attack needs to be robust for any weight initialization and any training stage of the shared global model. More formally,  the problem can be formulated as:
\begin{equation}
  \begin{aligned}
     \text{Constructing  } &(\mathbf{X}^{\prime}, \mathbf{Y}^{\prime})\\ \textrm{s.t.} \frac{\partial \mathcal{L}(\mathcal{F}(\mathbf{W},\mathbf{X}^{\prime}); \mathbf{Y}^{\prime})}{\partial \mathbf{W}} &= \frac{\partial \mathcal{L}(\mathcal{F}(\mathbf{W},\mathbf{X}); \mathbf{Y})}{\partial \mathbf{W}}
  \end{aligned}
\end{equation}
where $(\mathbf{X}^{\prime}, \mathbf{Y}^{\prime})$ are the reconstructed training data, i.e. images and labels for image tasks, and tokens and labels for language tasks, $\mathcal{L}(\mathcal{F}(\mathbf{W},\mathbf{X}^{\prime}); \mathbf{Y}^{\prime})$ is the loss for model $\mathcal{F}(\mathbf{W},\mathbf{X})$.

\section{Method And Analysis}
Now, we provide details on our SAPAG algorithm. 
\subsection{Dummy data and dummy gradients}
In order to reconstruct the training data, we first initialize the reconstructed training data as $(\mathbf{X}^\prime, \mathbf{Y}^\prime)$. We call $\mathbf{X}^\prime$ as the dummy input. $\mathbf{Y}^\prime$ is the dummy label. We can get the corresponding dummy gradient as
\begin{equation}
    \nabla{{\mathbf{W}}^\prime}  = \frac{\partial \mathcal{L}(\mathcal{F}(\mathbf{W},\mathbf {X}^\prime); {\mathbf{Y}}^\prime)}{\partial \mathbf{W}}
\end{equation}

\subsection{Distance between the dummy gradient and the ground truth gradient}
The next step in SAPAG is to optimize $\nabla{{\mathbf{W}}^\prime}$ and bring it closer to the ground truth gradient $\nabla{{\mathbf{W}}}$ as much as possible. In this case, we need to define a differentiable distance function $\mathcal{D}({\mathbf{W}},{\mathbf{W}^\prime})$, so that we can obtain the best ${\mathbf{X}^\prime}$ and ${\mathbf{Y}^\prime}$ as follows:

\begin{equation}
    ({\mathbf{X}^*},{\mathbf{Y}^*}) = \underset{ ({\bf{X}}^{\prime},{\bf{Y}}^{\prime}) }{\text{arg min }} {\mathcal{D}}(\nabla{{\mathbf{W}}^\prime},\nabla{\mathbf{W}})
\end{equation}

\subsection{Distance function for gradient matching}
We have observed empirically that in the weight initialization stage, the ground truth gradients $\nabla{{\mathbf{W}}}$ of the same training data are smaller when initializing the weights of the neural networks by a normal distribution than by a uniform distribution. Besides, the $\nabla{{\mathbf{W}}}$ under a normal weight initialization gathers around zero values more heavily than the $\nabla{{\mathbf{W}}}$ under a uniform weight initialization. When we obtain the dummy gradient $\nabla{{\mathbf{W}}^\prime}$ from the dummy data, we have noted that $\nabla{{\mathbf{W}}^\prime}$ values are much smaller than the $\nabla{{\mathbf{W}}}$ values under the same setting. If we use the Euclidean distance between $\nabla{{\mathbf{W}}^\prime}$ and $\nabla{{\mathbf{W}}}$ as the distance function, the reconstruction of the training data is driven by large gradients at the early stages. However, this might cause a problem under a normal weight initialization since most of the gradients gather around zero while a small proportion of gradients have large values.

To overcome this problem, instead of using the Euclidean distance between $\nabla{{\mathbf{W}}^\prime}$ and $\nabla{{\mathbf{W}}}$ as the distance function, we use a weighted Gaussian kernel based function as our distance function:
\begin{equation}
     \mathcal{D}(\nabla{{\mathbf{W}}}^{\prime},\nabla{{\mathbf{W}}}) = 
     Q\cdot (1-exp({\frac{{- \lVert \nabla{\mathbf{W}^{\prime}}-\nabla{\mathbf{W}} \rVert}^2} {\sigma^2}}))
\label{eq:lossw}
\end{equation}
where $Q$ is a factor specified for each layer's ${\nabla\mathbf{W}}$ and its value decreases along with the order of the layer. By doing this, we put larger weights on the gradient differences on the front layers as they are closer to the input training data. $\sigma^2(\nabla \mathbf{W})$ is the scaling factor associated with $\nabla \mathbf{W}$. The value of $\sigma^2$ is crucial and needs to be suitable for different weight settings. We have found that the gradient $\nabla{\mathbf{W}}$ roughly follows a Gaussian distribution with very long tails and centres at 0. We can use the variance of $\nabla{\mathbf{W}}$ to estimate the optimal $\sigma^2$:
\begin{equation}
    \sigma^2 = \textrm{Var}(\nabla\mathbf{W})
\label{eq:sigma}    
\end{equation}

Thus, it is adaptive to the attack target. 

Next, we will introduce the properties of our distance function in comparison with the Euclidean function of ${\nabla\mathbf{W}}$ and $\nabla\mathbf{W}^\prime$. The first derivative of the loss in Eq. \ref{eq:lossw} with respect to $\nabla\mathbf{W}^\prime$ can be written as:
\begin{equation}
    \frac{\partial \mathcal{D}} {\partial \nabla \mathbf{W}^\prime} = 2Q \cdot \frac{(\nabla\mathbf{W}^\prime-\nabla\mathbf{W})} {\sigma^2} exp({\frac{{-\lVert \nabla{\mathbf{W}^{\prime}}-\nabla{\mathbf{W}} \rVert}^2} {\sigma^2}})
\label{eq:drt1}    
\end{equation}

The second derivative of the loss in Eq. \ref{eq:lossw} with respect to $\nabla\mathbf{W}^\prime$ can be written as:
\begin{equation}
    \begin{aligned}
    \frac{\partial^2 \mathcal{D}} {\partial \nabla \mathbf{W}^{\prime 2}} =&
     4Q \cdot \frac{(\nabla\mathbf{W}^\prime-\nabla\mathbf{W})^2-\sigma^2} {\sigma^4} \\
    &\cdot exp({\frac{{-\lVert \nabla{\mathbf{W}^{\prime}}-\nabla{\mathbf{W}} \rVert}^2} {\sigma^2}})
    \end{aligned}    
\label{eq:drt12}
\end{equation}

We can infer from Eq. \ref{eq:drt12} that the max value of $\frac{\partial \mathcal{D}} {\partial \nabla \mathbf{W}^\prime}$ in Eq. \ref{eq:drt1} is  $2Q/(e\sigma)$ when $(\nabla\mathbf{W}^\prime-\nabla\mathbf{W})^2= \sigma^2$. The absolute magnitude of the derivative $\frac{\partial \mathcal{D}} {\partial \nabla \mathbf{W}^\prime}$ in Equation \ref{eq:drt1} first increases and then decreases along with an increase in the gradient difference between $\nabla\mathbf{W}^\prime$ and $\nabla\mathbf{W}$.

The first derivative of the Euclidean distance $\nabla\mathbf{W}^\prime$ and $\nabla\mathbf{W}$ with respect to $\nabla\mathbf{W}^\prime$ is:

\begin{equation}
    \frac{\partial \mathcal{D}} {\partial \nabla \mathbf{W}^\prime} = 2(\nabla \mathbf{W}^\prime-\nabla\mathbf{W})
\label{eq:drt2}    
\end{equation}

$\frac{\partial \mathcal{D}} {\partial \nabla \mathbf{W}^\prime}$ is in proportion with the magnitude of the gradient difference. Equation \ref{eq:drt1} can be viewed as a weighted sum of gradient differences comparing to equation \ref{eq:drt2}. The non-linearity of $\frac{\partial \mathcal{D}} {\partial \nabla \mathbf{W}^\prime}$ in Equation \ref{eq:drt1} can make it less affected by large gradients. The reconstruction process at an early stage for a normal weight initialization will be driven by the majority of the gradients instead of a small proportion of the large gradients.

\subsection{SAPAG algorithm}


Our complete SAPAG algorithm is shown in Algorithm 1, and the highlights of our SAPAG algorithm are as follows. 
We initialize a dummy data $(\mathbf{X}^\prime, \mathbf{Y}^\prime)$ and obtain the gradient $\nabla \mathbf{W}^\prime$ of the dummy data. We update the dummy data in each iteration to minimize the distance between the dummy gradient $\nabla \mathbf{W}^\prime$ and the real data gradient $\nabla \mathbf{W}$. In contrast to \cite{zhu2019deep}, we use a weighted Gaussian kernel based function in Eq.\ref{eq:lossw} as our distance function, and the initialization of the dummy data can be generated from a normal distribution or constant values $\mathbf{C}$ as shown in Algorithm \ref{alg:alg1}. At each step after updating $(\mathbf{X}^\prime, \mathbf{Y}^\prime)$ in line 5, we normalize the value of $(\mathbf{X}^\prime, \mathbf{Y}^\prime)$ to a certain range in line 6 to prevent it from being trapped at some extreme value and make the training more stable.
\begin{algorithm}[ht]
	\caption{The Process of SAPAG}
	\begin{algorithmic}[1]
	\State Initial: $\mathbf{X}^\prime \sim \mathcal{N}(0,1)$ or $C$,  $\mathbf{Y}^\prime \sim \mathcal{N}(0,1)$
		\For {$i$ in $Iterations$}
			\State Get gradient $\nabla\mathbf{W}^\prime \leftarrow \partial{\mathcal{L} (f(\mathbf{X},\mathbf{W})}/\partial{\mathbf{W}^\prime})$
			\State $\mathcal{D}(\nabla\mathbf{W}^\prime, \nabla\mathbf{W}_i)  \leftarrow  Q\cdot (1-exp({\frac{{- \lVert \nabla{\mathbf{W}^{\prime}}-\nabla{\mathbf{W}_i} \rVert}^2} {\sigma^2}}))$
			\State Update $(\mathbf{X}^\prime,\mathbf{Y}^\prime)$:
			\State
			  $\mathbf{X}^\prime \leftarrow \mathbf{X}^\prime-\eta \frac{\partial \mathcal{D(\nabla\mathbf{W}^\prime, \nabla\mathbf{W}_i)}} {\partial \nabla \mathbf{X}^\prime}$,
			\State
			  $\mathbf{Y}^\prime \leftarrow \mathbf{Y}^\prime-\eta \frac{\partial \mathcal{D(\nabla\mathbf{W}^\prime, \nabla\mathbf{W}_i)}} {\partial \nabla \mathbf{Y}^\prime}$

			\State Normalize $\mathbf{X}^\prime \in [\textrm{0, 1}]$
		\EndFor
	\end{algorithmic} 
	\label{alg:alg1}
\end{algorithm}
\subsection{Evaluation Metrics}
In order to evaluate the attack efficiency and the reconstruction quality, we use three different metrics to measure the difference/similarity between reconstructed training data and the real training data, namely Mean Square Error (MSE), Peak Signal-to-Noise Ratio (PSNR), and Structural Similarity Index Measure (SSIM).

MSE measures the difference between recovered data $\mathbf{X}^\prime$ and the training data $\mathbf{X}$ and is calculated as:

\begin{equation}
    MSE (\mathbf{X}^\prime, \mathbf{X}) = \sum_{i=1}^{M}\frac{1}{M}(\mathbf{X}^\prime_i-\mathbf{X}_i)^{2}
\label{eq:mse}    
\end{equation}
Where $\mathbf{X}_i$ and $\mathbf{X}_i^\prime$ are the $i$th pixel value in $\mathbf{X}$ and $\mathbf{X}^\prime$, $M$ is the total number of pixels.

PSNR is calculated as:

\begin{equation}
    PSNR (\mathbf{X}^\prime, \mathbf{X}) = 20\cdot \log_{10}(MAX)-10\cdot \log_{10}(MSE)
\label{eq:psnr}    
\end{equation}

where $MAX$ is the maximum possible pixel value of the image. 

SSIM~\cite{wang2004image} is a weighted combination of three comparison measures: luminance, contrast, and structure. When all the weights are 1, SSIM can be derived as:
\begin{equation}
    SSIM (\mathbf{X}^\prime, \mathbf{X}) = \frac{(2\mu_{\mathbf{X}^\prime}\mu_{\mathbf{X}}+c_1)(2\sigma_{\mathbf{X}^\prime\mathbf{X}}+c_2)}{(\mu_{\mathbf{X}^\prime}^2+\mu_{\mathbf{X}}^2+c_1)(\sigma_{\mathbf{X}^\prime}^2+\sigma_{\mathbf{X}}^2+c_2)}
\label{eq:ssim}    
\end{equation}
where $\mu_{\mathbf{X}^\prime}$ is the mean of $\mathbf{X}^\prime$, $\mu_{\mathbf{X}}$ is the mean of $\mathbf{X}$, $\sigma_{\mathbf{X}^\prime}$ is the variance of $\mathbf{X}^\prime$, $\sigma_{\mathbf{X}}$ is the variance of $\mathbf{X}$, $\sigma_{\mathbf{X}{\mathbf{X}^\prime}}$ is the covariance of $\mathbf{X}^\prime$ and $\mathbf{X}$. $c_1=(k_1L)^2$ and $c_2=(k_2L)^2$. $L$ is the dynamic range of the pixel-values, $k_1=0.01$ and $k_2=0.03$, by default.

\begin{figure}[ht]
\centering
	\includegraphics[width=0.45\textwidth]{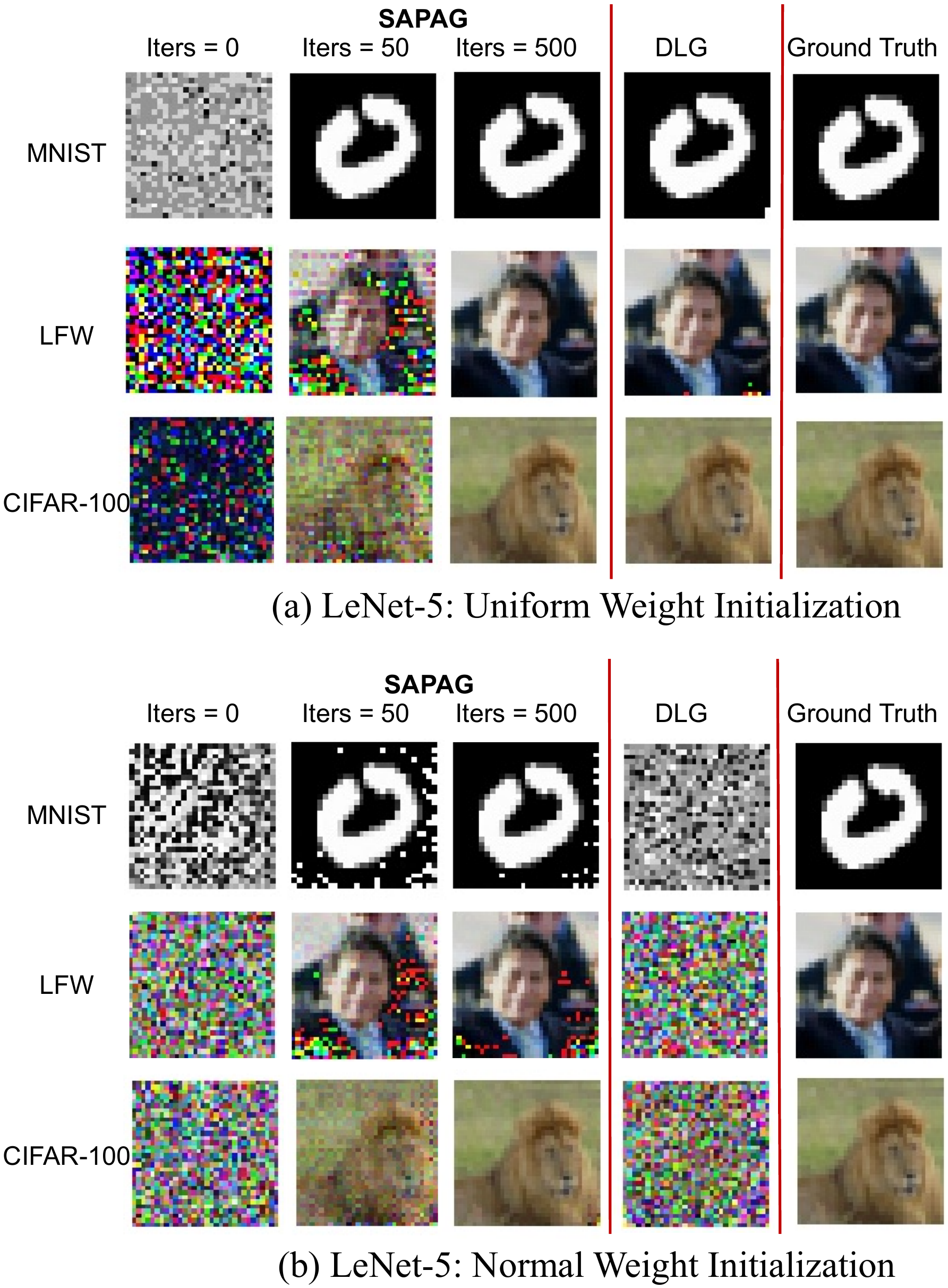}
    \vspace{-0.2 cm}
    \caption{ Reconstruction of images from MNIST, LFW and CIFAR-100 datasets for LeNet-5 with weights initialized by uniform and normal distributions. The DLG algorithm can only recover the training images under uniform weight initialization, while SAPAG can recover the training images under both uniform and normal weight initializations of LeNet-5.}
    \label{fig:Fig1}
\end{figure}
\begin{figure}[ht]
\centering
	\includegraphics[width=0.48\textwidth]{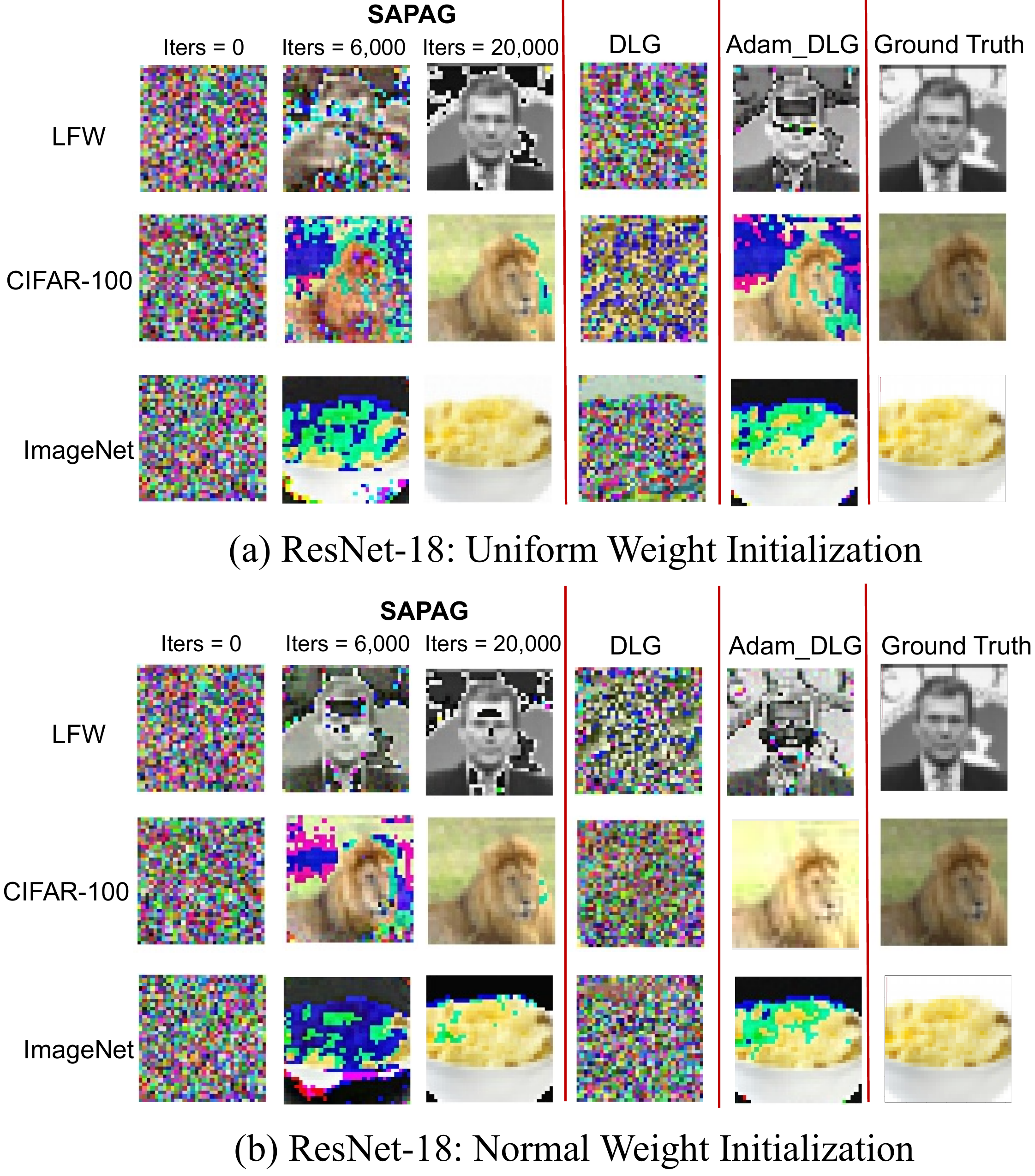}
    \vspace{-0.2 cm}
    \caption{ Reconstruction of images from LFW, CIFAR-100, and ImageNet datasets for ResNet18 with weights initialized by uniform and normal distributions. The optimizer used is L-BFGS for DLG and AdamW for the proposed method and Adam\_DLG. The proposed method achieves the best reconstruction quality under both uniform and normal weight initializations of ResNet-18.}
    \label{fig:Fig2}
\end{figure}
\vspace{-2mm}


\begin{figure}[ht]
\centering
	\includegraphics[width=0.45\textwidth]{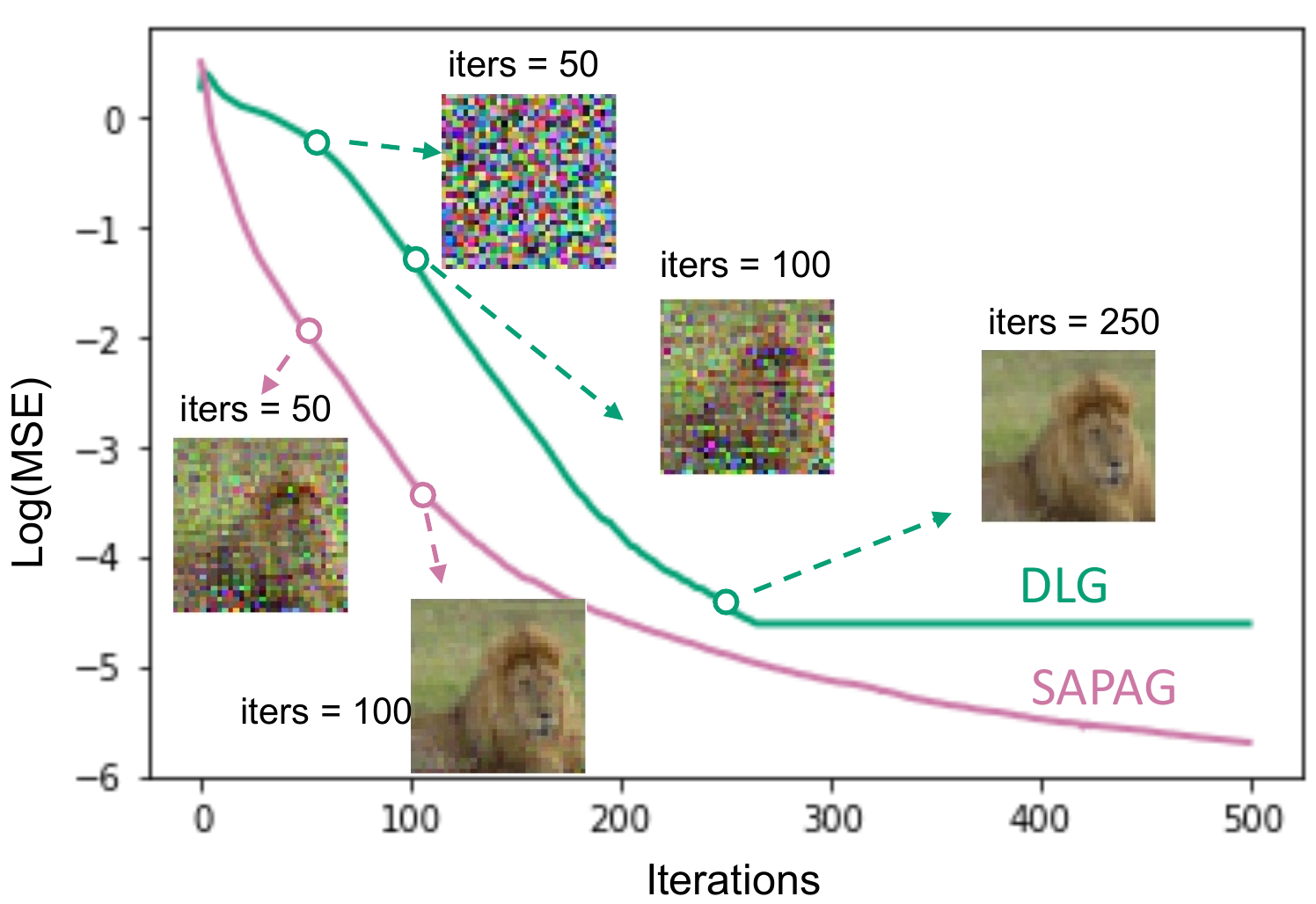}
    \vspace{-0.2 cm}
    \caption{The plot of MSE in logarithmic scale along with iterations for DLG and SAPAG. }
    \label{fig:train_speed}
\end{figure}

\section{Experimental Setup}
 All of our experiments are conducted on the AWS EC2 cloud instance with Intel(R) Xeon(R) Platinum 8175M (96 virtual CPUs with 748 GB memory) and 8 NVIDIA Tesla V100 GPUs (32 GB memory) and another server with Intel(R) Xeon(R) Gold 5218 (64 virtual CPUs with 504 GB memory) and 8 NVIDIA Quadro RTX 6000 GPUs (24GB memory) by PyTorch 1.5.1, Python 3.6, and CUDA 10.2.

\begin{table}[ht]
\centering
\scriptsize
\begin{subtable}[t]{0.45\textwidth}
\begin{tabular}{ccc|cc|cc}
\hline
& \multicolumn{2}{c|}{MNIST}                                                           
& \multicolumn{2}{c|}{CIFAR-100}
& \multicolumn{2}{c}{LFW}
\\ \hline
& \cellcolor[HTML]{FFFFFF}\begin{tabular}[c]{@{}c@{}}DLG  \end{tabular} & \cellcolor[HTML]{dddddd}\begin{tabular}[c]{@{}c@{}}SAPAG\end{tabular} & \begin{tabular}[c]{@{}c@{}}DLG \end{tabular} & \cellcolor[HTML]{dddddd}\begin{tabular}[c]{@{}c@{}}SAPAG\end{tabular} &
\begin{tabular}[c]{@{}c@{}}DLG \end{tabular} & \cellcolor[HTML]{dddddd}\begin{tabular}[c]{@{}c@{}}SAPAG\end{tabular}
\\ \hline
\begin{tabular}[c]{@{}c@{}}MSE\end{tabular}                          
        &1.39e-7 &\cellcolor[HTML]{dddddd}{\bf 1.50e-8}                                &4.21e-5 &\cellcolor[HTML]{dddddd}{\bf 2.57e-5} 
        &4.27e-5 &\cellcolor[HTML]{dddddd}{\bf 2.12e-6}                                
\\ \hline
\begin{tabular}[c]{@{}c@{}}PSNR\end{tabular}  
        &68.61 &\cellcolor[HTML]{dddddd}{\bf 78.23}                             
        &43.91 &\cellcolor[HTML]{dddddd}{\bf 55.91}  
        &43.76 &\cellcolor[HTML]{dddddd}{\bf 56.74}                      
\\ \hline
\begin{tabular}[c]{@{}c@{}}SSIM\end{tabular}  
        &{\bf 1.00} &\cellcolor[HTML]{dddddd}{\bf 1.00}                              
        &{\bf 1.00} &\cellcolor[HTML]{dddddd}{\bf 1.00}  
        &{\bf 1.00} &\cellcolor[HTML]{dddddd}{\bf 1.00} 
\\ \hline

\end{tabular}
\caption{LeNet-5: Uniform Weight Initialization}
\end{subtable}

\begin{subtable}[t]{0.45\textwidth}
\begin{tabular}{ccc|cc|cc}
\hline
& \multicolumn{2}{c|}{MNIST}                                                           
& \multicolumn{2}{c|}{CIFAR-100}
& \multicolumn{2}{c}{LFW}
\\ \hline
& \cellcolor[HTML]{FFFFFF}\begin{tabular}[c]{@{}c@{}}DLG  \end{tabular} & \cellcolor[HTML]{dddddd}\begin{tabular}[c]{@{}c@{}}SAPAG\end{tabular} & \begin{tabular}[c]{@{}c@{}}DLG \end{tabular} & \cellcolor[HTML]{dddddd}\begin{tabular}[c]{@{}c@{}}SAPAG\end{tabular} &
\begin{tabular}[c]{@{}c@{}}DLG \end{tabular} & \cellcolor[HTML]{dddddd}\begin{tabular}[c]{@{}c@{}}SAPAG\end{tabular}
\\ \hline
\begin{tabular}[c]{@{}c@{}}MSE\end{tabular}                          
        &1.10 &\cellcolor[HTML]{dddddd}{\bf 2.77e-5}                               &0.64 &\cellcolor[HTML]{dddddd}{\bf 5.76e-4}  
        &1.27 &\cellcolor[HTML]{dddddd}{\bf 8.4e-4}                                
\\ \hline
\begin{tabular}[c]{@{}c@{}}PSNR\end{tabular}  
        &-0.41 &\cellcolor[HTML]{dddddd}{\bf 45.60}                              
        &3.75 &\cellcolor[HTML]{dddddd}{\bf 32.43} 
        &-1.04 &\cellcolor[HTML]{dddddd}{\bf 30.72}                      
\\ \hline
\begin{tabular}[c]{@{}c@{}}SSIM\end{tabular}  
        &3.20e-3 &\cellcolor[HTML]{dddddd}{\bf 0.99}                              
        &7.90e-2 &\cellcolor[HTML]{dddddd}{\bf 0.98} 
        &7.55e-3 &\cellcolor[HTML]{dddddd}{\bf 0.98}   
\\ \hline

\end{tabular}
\caption{LeNet-5: Normal Weight Initialization}
\end{subtable}

\caption{A comparison of the reconstruction quality of  MNIST, CIFAR-100 and LFW for LeNet-5 between the DLG and the proposed method.  }
\label{table:rec_len}
\end{table}


\begin{table*}[ht]
\centering
\scriptsize
\begin{subtable}[t]{1\textwidth}
\centering
{\begin{tabular}{cccc|ccc|ccc}
\hline
& \multicolumn{3}{c|}{CIFAR-100}                                                           
& \multicolumn{3}{c|}{LFW}
& \multicolumn{3}{c}{ImageNet}
\\ \hline
& \cellcolor[HTML]{FFFFFF}\begin{tabular}[c]{@{}c@{}}DLG  \end{tabular} &
  \begin{tabular}[c]{@{}c@{}}Adam\_DLG  \end{tabular} &
  \cellcolor[HTML]{dddddd}\begin{tabular}[c]{@{}c@{}}SAPAG\end{tabular} &
  \begin{tabular}[c]{@{}c@{}}DLG \end{tabular} &
  \begin{tabular}[c]{@{}c@{}}Adam\_DLG \end{tabular} &
  \cellcolor[HTML]{dddddd}\begin{tabular}[c]{@{}c@{}}SAPAG\end{tabular} &
\begin{tabular}[c]{@{}c@{}}DLG \end{tabular} &
\begin{tabular}[c]{@{}c@{}}Adam\_DLG \end{tabular} &
\cellcolor[HTML]{dddddd}\begin{tabular}[c]{@{}c@{}}SAPAG\end{tabular}
\\ \hline
\begin{tabular}[c]{@{}c@{}}MSE\end{tabular}                          
        &510.91 &0.14&\cellcolor[HTML]{dddddd}{\bf 0.03}                               
        &347.79 &0.18&\cellcolor[HTML]{dddddd}{\bf 0.06}  
        &1535.97 &0.18&\cellcolor[HTML]{dddddd}{\bf 0.04}                                
\\ \hline
\begin{tabular}[c]{@{}c@{}}PSNR\end{tabular}  
        &-27.08 &8.46&\cellcolor[HTML]{dddddd}{\bf 15.16}                               
        &-25.41 &7.27&\cellcolor[HTML]{dddddd}{\bf 12.27}  
        &-31.84 &7.49&\cellcolor[HTML]{dddddd}{\bf 14.48}                     
\\ \hline
\begin{tabular}[c]{@{}c@{}}SSIM\end{tabular}  
        &1.63e-3 &0.72&\cellcolor[HTML]{dddddd}{\bf 0.86}                               
        &1.69e-3 &0.67&\cellcolor[HTML]{dddddd} {\bf 0.83} 
        &7.86e-4 &0.65&\cellcolor[HTML]{dddddd}{\bf 0.93}
\\ \hline

\end{tabular}}
\caption{ResNet-18: Uniform Weight Initialization}
\end{subtable}

\begin{subtable}[t]{1\textwidth}
\centering
{\begin{tabular}{cccc|ccc|ccc}
\hline
& \multicolumn{3}{c|}{CIFAR-100}                                                           
& \multicolumn{3}{c|}{LFW}
& \multicolumn{3}{c}{ImageNet}
\\ \hline
& \cellcolor[HTML]{FFFFFF}\begin{tabular}[c]{@{}c@{}}DLG  \end{tabular} &
  \begin{tabular}[c]{@{}c@{}}Adam\_DLG  \end{tabular} &
  \cellcolor[HTML]{dddddd}\begin{tabular}[c]{@{}c@{}}SAPAG\end{tabular} &
  \begin{tabular}[c]{@{}c@{}}DLG \end{tabular} &
  \begin{tabular}[c]{@{}c@{}}Adam\_DLG \end{tabular} &
  \cellcolor[HTML]{dddddd}\begin{tabular}[c]{@{}c@{}}SAPAG\end{tabular} &
\begin{tabular}[c]{@{}c@{}}DLG \end{tabular} &
\begin{tabular}[c]{@{}c@{}}Adam\_DLG \end{tabular} &
\cellcolor[HTML]{dddddd}\begin{tabular}[c]{@{}c@{}}SAPAG\end{tabular}
\\ \hline
\begin{tabular}[c]{@{}c@{}}MSE\end{tabular}                          
        &1.16 &0.12&\cellcolor[HTML]{dddddd}{\bf 0.01}                                
        &1.31 &0.27&\cellcolor[HTML]{dddddd} {\bf 0.04} 
        &8126.61 &0.28&\cellcolor[HTML]{dddddd}{\bf 0.05}                               
\\ \hline
\begin{tabular}[c]{@{}c@{}}PSNR\end{tabular}  
        &-0.66&9.37 &\cellcolor[HTML]{dddddd}{\bf 19.02}                                
        &-1.13 &5.63&\cellcolor[HTML]{dddddd}{\bf 14.04}
        &-36.60 &5.47&\cellcolor[HTML]{dddddd}{\bf 12.90}                      
\\ \hline
\begin{tabular}[c]{@{}c@{}}SSIM\end{tabular}  
        &1.22e-2 &0.76 &\cellcolor[HTML]{dddddd}{\bf 0.95}                               
        &5.12e-3 &0.64 &\cellcolor[HTML]{dddddd} {\bf 0.87}
        &1.79e-4 &0.56 &\cellcolor[HTML]{dddddd}{\bf 0.89}  
\\ \hline

\end{tabular}}
\caption{ResNet-18: Normal Weight Initialization}
\end{subtable}
\caption{A comparison of the reconstruction quality of CIFAR-100, LFW, and ImageNet for ResNet-18 between the DLG and the proposed method.  }
\label{table:rec_res}
\end{table*}


\section{Evaluation on Computer Vision Tasks}
 \subsection{Datasets and Networks}
To make a general evaluation, we use different image datasets including MNIST~\cite{deng2012mnist}, CIFAR-100~\cite{krizhevsky2009learning}, LFW~\cite{huang2008labeled} and ImageNet~\cite{deng2009imagenet}. MNIST, CIFAR-100, LFW, and ImageNet have 10, 100, 5,749, and 1,000 classes, respectively.  We also consider different DNNs with different levels of depth, including LeNet-5 and ResNet-18. LeNet-5 consists of 4 convolutional layers and one fully connected layer and uses Sigmoid as the activation function. The kernel size, the number of output channels, and the strides for each convolutional layer are 5, 12, and 1, respectively. ResNet-18 consists of 17 convolutional layers and one fully connected layer and uses Sigmoid as the activation function. The kernel size of each convolutional layer is 3. The number of output channels for the first convolutional layer is 64. The numbers of output channels for the following four blocks are 64, 128, 256, and 512. The stride for all the convolutional layers is 1.

\begin{figure}[t]
\centering
	\includegraphics[width=0.45\textwidth]{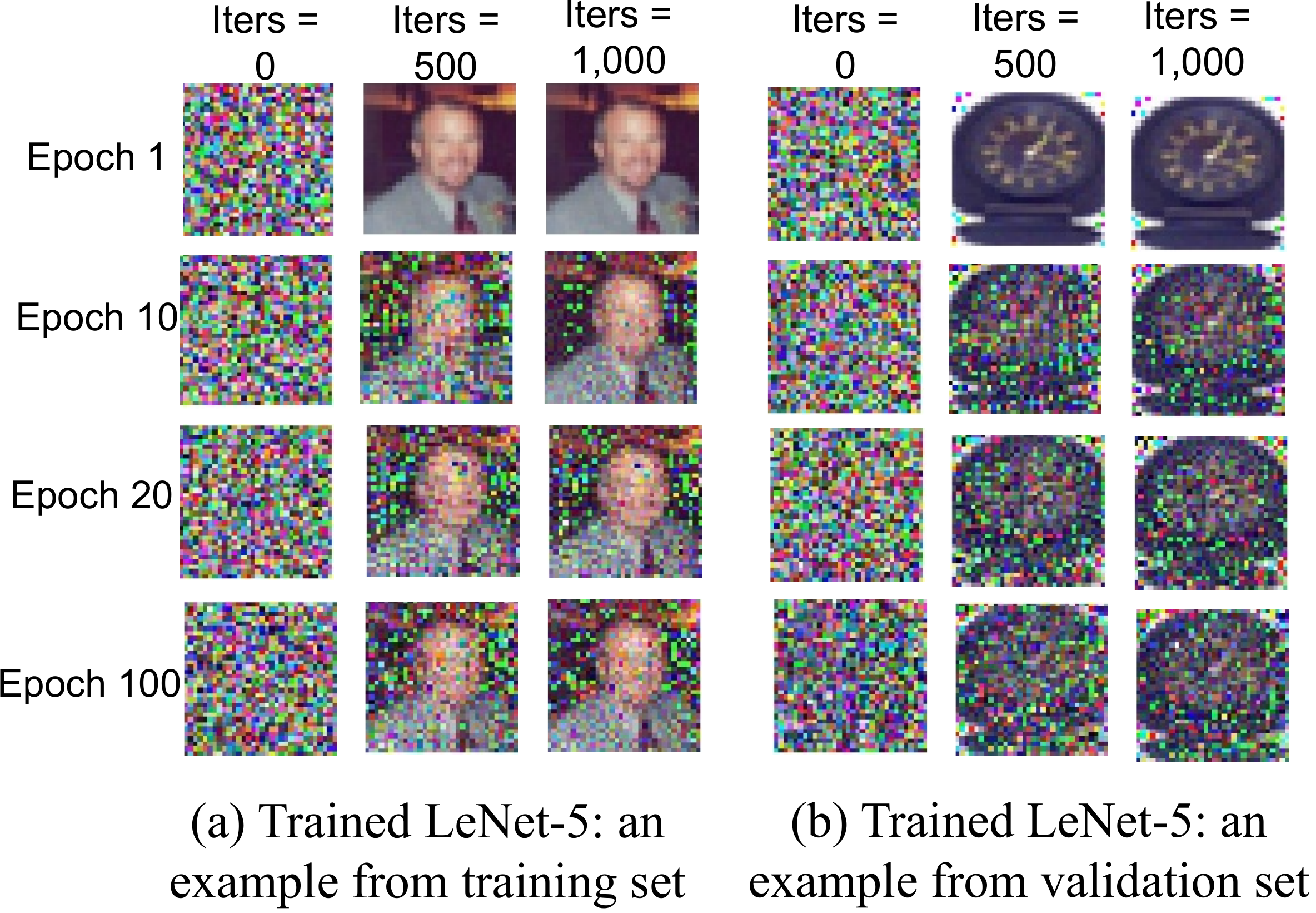}
    \vspace{-0.2 cm}
    \caption{ Reconstruction of images from the training set and validation set for LeNet-5 trained for 1, 5, and 10 epochs.}
    \label{fig:trained}
\end{figure}

\begin{figure}[ht]
\centering
	\includegraphics[width=0.48\textwidth]{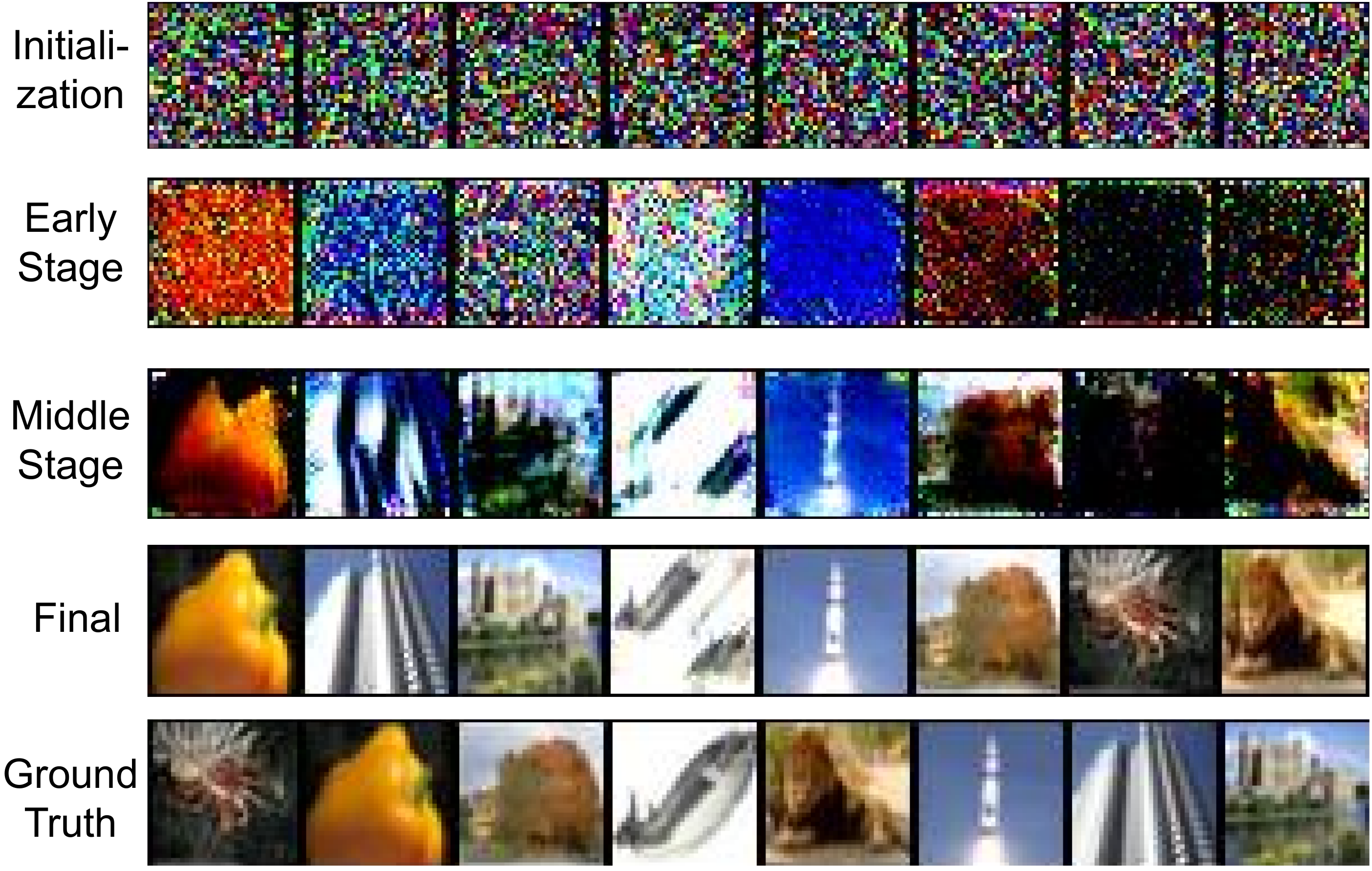}
    \vspace{-0.2 cm}
    \caption{ SAPAG reconstruction of a batch of 8 images from CIFAR-100 dataset for ResNet-18 with weights initialized by a normal distribution.}
    \label{fig:batch}
\end{figure}

\subsection{Results for Untrained Networks} 

To evaluate the performance of SAPAG, we first apply it to untrained networks with a uniform weight initialization and Xavier normal weight initialization~\cite{glorot2010understanding}. The range of the uniform distribution is (-0.5, 0.5), and the gain factor of the Xavier normal distribution is 1. The optimizer used in experiments on LeNet-5 is L-BFGS~\cite{liu1989limited} for both and the optimizer used in experiments on ResNet-18 is AdamW~\cite{loshchilov2017decoupled}.  The learning rate we have used is 1 for the L-BFGS optimizer and 0.001 for the AdamW optimizer. The maximum number of iterations is set as 500 for the L-BFGS optimizer and 20,000 for the AdamW optimizer.

The reconstruction quality of MNIST, LFW, and CIFAR-100, on LeNet-5, are shown in Table~\ref{table:rec_len} and  LFW, CIFAR-100, and ImageNet on ResNet-18 are shown in Table~\ref{table:rec_res}. The results indicate that the proposed attack method generally performs better than the DLG method in terms of reconstruction quality measures MSE, PSNR, and SSIM. Both our method and DLG have the best reconstruction quality on MNIST dataset, which is intuitive since MNIST dataset has less complicated patterns than CIFAR-100 and LFW. Figure~\ref{fig:Fig1} and~\ref{fig:Fig2} show examples from different datasets comparing with DLG and the ground truth. For LeNet-5, our method can recover the major shape of an image after only 50 iterations and converge after 500 iterations using the L-BFGS optimizer. And for ReNset-18, our method can recover the major shape of an image after only 6,000 iterations and converge after 20,000 iterations for the AdamW optimizer. Figure \ref{fig:train_speed} compares the convergence speed of SAPAG and DLG for reconstructing an image from CIFAR-100 for LeNet-5. It shows that SAPAG has a faster convergence speed and converges at a smaller MSE than DLG.

\subsubsection{The impact of weight initialization}
A comparison of the reconstruction quality between uniform and normal weight initializations in Table ~\ref{table:rec_len} and ~\ref{table:rec_res} show that our method has a slightly better reconstruction for uniform weight initialization than normal weight initialization on LeNet-5, while reverse is true on ResNet-18. Nevertheless, our method can have a reasonably good reconstruction on both weight initialization settings. However, we have found that DLG can only recover images under a uniform weight initialization on LeNet-5 using the L-BFGS optimizer and ResNet-18 using the AdamW optimizer.       
\subsubsection{The impact of dummy image initialization}
We have realized that the initialization of dummy images is also crucial to the reconstruction of the training data. We have employed two different initialization strategies: random initialization and constant initialization. For some attack targets, the random initialization of dummy image can be faster and yield a better reconstruction of the training image than the constant initialization of dummy images, while the reverse is true for some other attack targets. It depends on the unique pattern of each attack target. 
\subsubsection{The impact of optimizer}
Our experimental results reveal that the L-BFGS optimizer speeds up convergence when optimizing dummy images on LeNet-5 but has the problem of unstable training. For ResNet-18, the L-BFGS optimizer usually converges at a point where the loss is not small enough and the training image is not well recovered. However, optimizer AdamW is more stable and can achieve much smaller loss upon convergence. A comparison of $L_2$ norm loss using the L-BFGS and AdamW optimizers in Table ~\ref{table:rec_res} shows that AdamW optimizer has a much smaller MSE and a higher PSNR and SSIM than the L-BFGS optimizer. 
\subsubsection{The impact of network depth}
LeNet-5 has 5 layers in total and ResNet-18 has 18 layers in total and 183 times more trainable parameters than LeNet-5. The reconstruction quality of CIFAR-100 and LFW datasets on ResNet-18 is lower than that on LeNet-5 in terms of all the three evaluation metrics. However, as a deeper network, ResNet-18 is not able to protect privacy either. The decrease of PSNR or SSIM mainly comes from noise in the background and luminance differences of images. As shown in Figure~\ref{fig:Fig2}, the main structure of images is leaked.

\begin{table*}[ht]
\centering

\label{if_rba312ex}
\begin{tabular}{|l|l|l|r|r|r|r|r|r|r|}
\hline
\multicolumn{1}{|l|}{} & \multicolumn{1}{l|}{Example 1} & \multicolumn{1}{l|}{Example 2}\\ \hline

SAPAG                                
& \begin{tabular}[c]{@{}l@{}}{\bf nonstudent advisor or} or {\bf collaborators} 
\\ collaborators {\bf should} appropriately as {\bf appropriately} 
\\ {\bf coauthors or} or be{\bf. however}\end{tabular}
& \begin{tabular}[c]{@{}l@{}}{\bf students are requested to honor} 
\\ {\bf the} submitting {\bf spirit} student by {\bf by}
\\ {\bf submitting} references of {\bf for}\end{tabular}
\\ \hline
DLG                                
& \begin{tabular}[c]{@{}l@{}}{\bf nonstudent} investigators {\bf advisors} as, 
\\ {\bf collaborators should} collaborators primary will
\\ September will will session work \end{tabular}                              
& \begin{tabular}[c]{@{}l@{}} {\bf students} to students been the the  
\\ collaborators the conference the accepted the 
\\ {\bf submitting} no the September will will session {\bf work} \end{tabular}
\\ \hline
Ground Truth                                
& \begin{tabular}[c]{@{}l@{}}nonstudent advisor or collaborators 
\\ should be acknowledged appropriately, 
\\ as coauthors or otherwise. however \end{tabular}                              
& \begin{tabular}[c]{@{}l@{}}students are requested to honor 
\\ the spirit of the program by 
\\ submitting only work for \end{tabular}
\\ \hline

\end{tabular}
\caption{Examples of privacy attack from gradients on language model.}
\label{tab:nlp}
\end{table*}

\subsection{Results for Trained Networks}
Hypothetically, the attack can happen at any time during the local training process. Except for weight initialization, we also evaluate SAPAG on networks trained for different numbers of epochs. We aim to explore how the change in weight distribution can affect the attack effectiveness. The experiment is applied on CIFAR-100 dataset and on LeNet-5 network. The optimizer used is L-BFGS, the learning rate is 1, and the maximum number of iterations is 500. To study the impact on training data and validation data, we attack images from both training data and validation data on LeNet-5 networks trained for 1, 5, and 10 epochs.

Figure \ref{fig:trained} shows the reconstruction of an example from the training set and validation set of CIFAR-100 on trained LeNet-5 network. In the early stages of the training process of LeNet-5 (epoch = 1), the attack can recover the training image and validation image with very few noises. After 10 epochs, the noises in the reconstructed images increases, but the objects in the images are recognizable.  When the training epoch of LeNet-5 reaches 100, the reconstructed images are still recognizable. We found that the construction quality of the training images (used to train the model) and validation images (not used to train the model) has no significant difference. We found empirically that the training of networks would cause the gradients of the training data decreases, thus the gradients carries less information. The scaling factor in the Gaussian kernel in SAPAG is adaptive to the gradient distribution of the attack target. Therefore, SAPAG is able to reconstruct the training image even when its gradients become very small.
\vspace{-2mm}

\subsection{Results for Batched Training Data}
Now we have demonstrated that the proposed attack works well on one single image under different network settings. In the practice of distributed training, the gradients shared come from a batch of training data instead of one single training data. Except for evaluating our work on a single pair of input and labels, we also evaluate our proposed attack method on the data where the batch size is larger than 1.  We randomly sample a batch of 8 images from CIFAR-100 and get their gradients from ResNet-18 with weight initialized by a normal distribution. The architecture of ResNet-18 is the same as above. We then randomly initialize 8 dummy images and stack them together and get their gradients similarly. The optimizer used is AdamW and the learning rate is set as 0.001.

Figure ~\ref{fig:batch} shows the results of recovering a random batch of training data including 8 images.  At the early stage of reconstruction, the main color of each image was first recovered for images 1, 2, 4, 5, and 6. In the middle stages of reconstruction, the shape of each image was then recovered. Finally, the recovered batch images are very similar to the ground truth in terms of both color and shape but in a different order,  except that the wale in the fifth image was cut into halves in the recovered image. 

We found that the reconstruction of batch training data does not require additional training time. The reconstruction process for the 8 images converges after 20,000 iterations using AdamW optimizer. The reconstruction speed is similar to that for one single image. Besides, the reconstruction quality does not decrease for the batch training data.


\section{Evaluation on Natural Language Processing Tasks}

The data we use are some random texts from the website page. The Network consists of one position encoder, two transformer encoders, and one transformer decoder~\cite{vaswani2017attention}. The number of heads in the self-attention layer is 4 and The dimension of the feed-forward network model is 200. The activation function is Gaussian Error Linear Units (GELU)\cite{hendrycks2016gaussian}.

We embed the text tokens into an embedding space of 5,000 dimensions and feed the embedding to the Transformer model. The embedding weights are uniformly initialized. The dummy data is a randomly initialized dummy embedding. The optimizer we used in the attack method is AdamW and the maximum number of iterations is set as 20,000. When we obtain the reconstructed embedding from SAPAG, tokens can be recovered as the multiplication of reconstructed embedding matrix and Moore–Penrose pseudoinverse matrix~\cite{barata2012moore} of the weights matrix. Finally, we get the reconstructed sentence from the tokens via vocabulary. Table \ref{tab:nlp} compares the reconstruction of two sentences including 15 words from website page for SAPAG and DLG. The words that match the ground truth text are highlighted in bold. SAPAG is able to recover most key words in the ground truth text, while DLG catches fewer words in the ground truth text. In addition, the results of SAPAG are closer to continuous and meaningful sentences.


\section{Conclusions}
In this paper, we provide a privacy attack from gradients: SAPAG. We demonstrate that SAPAG can successfully reconstruct the training data for different DNNs with different weight initializations and for DNNs in any training phases. Our experiments show that SAPAG has a faster convergence speed and higher reconstruction accuracy than the DLG algorithm. The experiments on a transformer-based language model show that SAPAG can successfully reconstruct token-wise training text. 
We found that L-BFGS optimizer works well when recovering the training data for LeNet-5 but it is not stable on ResNet-18. AdamW optimizer is more stable and has a better performance on ResNet-18. For the same training data, SAPAG has lower reconstruction accuracies from gradients on a deeper neural network (ResNet-18) than on a shallow neural network (LeNet-5), but the reconstruction is still efficient for ResNet-18. 
We believe that our work provides an important guide towards secure distributed learning.

\bibliographystyle{splncs}
\bibliography{egbibai}

\begin{thebibliography}{10}

\bibitem{chilimbi2014project}
Chilimbi, T., Suzue, Y., Apacible, J., Kalyanaraman, K.:
\newblock Project adam: Building an efficient and scalable deep learning
  training system.
\newblock In: 11th $\{$USENIX$\}$ Symposium on Operating Systems Design and
  Implementation ($\{$OSDI$\}$ 14). (2014)  571--582

\bibitem{shokri2015privacy}
Shokri, R., Shmatikov, V.:
\newblock Privacy-preserving deep learning.
\newblock In: Proceedings of the 22nd ACM SIGSAC conference on computer and
  communications security. (2015)  1310--1321

\bibitem{moritz2015sparknet}
Moritz, P., Nishihara, R., Stoica, I., Jordan, M.I.:
\newblock Sparknet: Training deep networks in spark.
\newblock arXiv preprint arXiv:1511.06051 (2015)

\bibitem{zinkevich2010parallelized}
Zinkevich, M., Weimer, M., Li, L., Smola, A.J.:
\newblock Parallelized stochastic gradient descent.
\newblock In: Advances in neural information processing systems. (2010)
  2595--2603

\bibitem{jochems2016distributed}
Jochems, A., Deist, T.M., Van~Soest, J., Eble, M., Bulens, P., Coucke, P.,
  Dries, W., Lambin, P., Dekker, A.:
\newblock Distributed learning: developing a predictive model based on data
  from multiple hospitals without data leaving the hospital--a real life proof
  of concept.
\newblock Radiotherapy and Oncology \textbf{121}(3) (2016)  459--467

\bibitem{fredrikson2015model}
Fredrikson, M., Jha, S., Ristenpart, T.:
\newblock Model inversion attacks that exploit confidence information and basic
  countermeasures.
\newblock In: Proceedings of the 22nd ACM SIGSAC Conference on Computer and
  Communications Security. (2015)  1322--1333

\bibitem{hitaj2017deep}
Hitaj, B., Ateniese, G., Perez-Cruz, F.:
\newblock Deep models under the gan: information leakage from collaborative
  deep learning.
\newblock In: Proceedings of the 2017 ACM SIGSAC Conference on Computer and
  Communications Security. (2017)  603--618

\bibitem{melis2019exploiting}
Melis, L., Song, C., De~Cristofaro, E., Shmatikov, V.:
\newblock Exploiting unintended feature leakage in collaborative learning.
\newblock In: 2019 IEEE Symposium on Security and Privacy (SP), IEEE (2019)
  691--706

\bibitem{zhu2019deep}
Zhu, L., Liu, Z., Han, S.:
\newblock Deep leakage from gradients.
\newblock In: Advances in Neural Information Processing Systems. (2019)
  14774--14784

\bibitem{he2016deep}
He, K., Zhang, X., Ren, S., Sun, J.:
\newblock Deep residual learning for image recognition.
\newblock In: Proceedings of the IEEE conference on computer vision and pattern
  recognition. (2016)  770--778

\bibitem{vaswani2017attention}
Vaswani, A., Shazeer, N., Parmar, N., Uszkoreit, J., Jones, L., Gomez, A.N.,
  Kaiser, {\L}., Polosukhin, I.:
\newblock Attention is all you need.
\newblock In: Advances in neural information processing systems. (2017)
  5998--6008

\bibitem{akiba2017chainermn}
Akiba, T., Fukuda, K., Suzuki, S.:
\newblock Chainermn: Scalable distributed deep learning framework.
\newblock arXiv preprint arXiv:1710.11351 (2017)

\bibitem{dean2012large}
Dean, J., Corrado, G., Monga, R., Chen, K., Devin, M., Mao, M., Ranzato, M.,
  Senior, A., Tucker, P., Yang, K.,  et~al.:
\newblock Large scale distributed deep networks.
\newblock In: Advances in neural information processing systems. (2012)
  1223--1231

\bibitem{mcmahan2017federated}
McMahan, B., Ramage, D.:
\newblock Federated learning: Collaborative machine learning without
  centralized training data.
\newblock Google Research Blog (2017)

\bibitem{bonawitz2019towards}
Bonawitz, K., Eichner, H., Grieskamp, W., Huba, D., Ingerman, A., Ivanov, V.,
  Kiddon, C., Konecny, J., Mazzocchi, S., McMahan, H.B.,  et~al.:
\newblock Towards federated learning at scale: System design.
\newblock arXiv preprint arXiv:1902.01046 (2019)

\bibitem{hard2018federated}
Hard, A., Rao, K., Mathews, R., Ramaswamy, S., Beaufays, F., Augenstein, S.,
  Eichner, H., Kiddon, C., Ramage, D.:
\newblock Federated learning for mobile keyboard prediction.
\newblock arXiv preprint arXiv:1811.03604 (2018)

\bibitem{goodfellow2014generative}
Goodfellow, I., Pouget-Abadie, J., Mirza, M., Xu, B., Warde-Farley, D., Ozair,
  S., Courville, A., Bengio, Y.:
\newblock Generative adversarial nets.
\newblock In: Advances in neural information processing systems. (2014)
  2672--2680

\bibitem{wang2004image}
Wang, Z., Bovik, A.C., Sheikh, H.R., Simoncelli, E.P.:
\newblock Image quality assessment: from error visibility to structural
  similarity.
\newblock IEEE transactions on image processing \textbf{13}(4) (2004)  600--612

\bibitem{deng2012mnist}
Deng, L.:
\newblock The mnist database of handwritten digit images for machine learning
  research [best of the web].
\newblock IEEE Signal Processing Magazine \textbf{29}(6) (2012)  141--142

\bibitem{krizhevsky2009learning}
Krizhevsky, A., Hinton, G.,  et~al.:
\newblock Learning multiple layers of features from tiny images.
\newblock (2009)

\bibitem{huang2008labeled}
Huang, G.B., Mattar, M., Berg, T., Learned-Miller, E.:
\newblock Labeled faces in the wild: A database forstudying face recognition in
  unconstrained environments.
\newblock (2008)

\bibitem{deng2009imagenet}
Deng, J., Dong, W., Socher, R., Li, L.J., Li, K., Fei-Fei, L.:
\newblock Imagenet: A large-scale hierarchical image database.
\newblock In: 2009 IEEE conference on computer vision and pattern recognition,
  Ieee (2009)  248--255

\bibitem{glorot2010understanding}
Glorot, X., Bengio, Y.:
\newblock Understanding the difficulty of training deep feedforward neural
  networks.
\newblock In: Proceedings of the thirteenth international conference on
  artificial intelligence and statistics. (2010)  249--256

\bibitem{liu1989limited}
Liu, D.C., Nocedal, J.:
\newblock On the limited memory bfgs method for large scale optimization.
\newblock Mathematical programming \textbf{45}(1-3) (1989)  503--528

\bibitem{loshchilov2017decoupled}
Loshchilov, I., Hutter, F.:
\newblock Decoupled weight decay regularization.
\newblock arXiv preprint arXiv:1711.05101 (2017)

\bibitem{hendrycks2016gaussian}
Hendrycks, D., Gimpel, K.:
\newblock Gaussian error linear units (gelus).
\newblock arXiv preprint arXiv:1606.08415 (2016)

\bibitem{barata2012moore}
Barata, J.C.A., Hussein, M.S.:
\newblock The moore--penrose pseudoinverse: A tutorial review of the theory.
\newblock Brazilian Journal of Physics \textbf{42}(1-2) (2012)  146--165

\end{thebibliography}
\end{document}